\documentclass[conference]{IEEEtran}
\IEEEoverridecommandlockouts
\usepackage{cite}
\usepackage{amsmath,amssymb,amsfonts,amsthm}
\usepackage{algorithmic}
\usepackage{graphicx}
\usepackage{textcomp}
\usepackage{xcolor}
\def\BibTeX{{\rm B\kern-.05em{\sc i\kern-.025em b}\kern-.08em
    T\kern-.1667em\lower.7ex\hbox{E}\kern-.125emX}}

\theoremstyle{definition}
\newtheorem{theorem}{Theorem}[section]

\newtheorem{example}[theorem]{Example}

\def \x{\mathbf{x}}

\def \x {\mathbf{x}}

\def \D {\mathbf{D}}

\def \F {\mathbf{F}}

\def \H{\mathbf{H}}
\def \I{\mathbf{H}}

\def \M {\mathbf{M}}

\newcommand{\R}{\mathbf{R}}

\def \V {\mathbf{V}}
\def \X {\mathbf{X}}

\newcommand{\one}{\mathbf{1}}
\newcommand{\Lam}{\mathbf{\Lambda}}

\begin{document}

\title{Multi-Dimensional Scaling on Groups\\
\thanks{ $\dagger$ This author was supported by the National Science Foundation under contract  CCF-1712788}
}

\author{\IEEEauthorblockN{Mark Blumstein$^\dagger$}
\IEEEauthorblockA{\textit{Department of Mathematics} \\
\textit{Colorado State University}\\
Fort Collins, USA \\
blumstei@math.colostate.edu}
\and
\IEEEauthorblockN{Henry Kvinge}
\IEEEauthorblockA{\textit{Pacific Northwest National Laboratory}\\
\textit{Seattle Research Center}\\
Seattle, WA, USA \\
henry.kvinge@pnnl.gov}
}

\maketitle

\begin{abstract}
Leveraging the intrinsic symmetries in data for clear and efficient analysis is an important theme in signal processing and other data-driven sciences. A basic example of this is the ubiquity of the discrete Fourier transform which arises from translational symmetry (i.e. time-delay/phase-shift). 

Particularly important in this area is understanding how symmetries inform the algorithms that we apply to our data. In this paper we explore the behavior of the dimensionality reduction algorithm multi-dimensional scaling (MDS) in the presence of symmetry. We show that understanding the properties of the underlying symmetry group allows us to make strong statements about the output of MDS even before applying the algorithm itself. In analogy to Fourier theory, we show that in some cases only a handful of fundamental ``frequencies'' (irreducible representations derived from the corresponding group) contribute information for the MDS Euclidean embedding.
\end{abstract}


\section{Introduction}


The use of groups and representation theory in the data-driven sciences has a long if understated history. The canonical reference is Diaconis' book \cite{Dia88} which shows the surprisingly broad range of real world problems in statistics and probability that can be solved by utilizing tools from representation theory. 

More recently, convolutional neural networks have made remarkable strides toward solving problems in computer vision by utilizing the group convolution for $\mathbb{Z}\times \mathbb{Z}$ to achieve invariance to translation in images \cite{krizhevsky2012imagenet}. researchers in machine learning have been exploring how symmetries and their corresponding groups can be built into machine learning algorithms to achieve invariance to certain types of structured variation. Some recent examples include \cite{cohen2016group,cohen2017convolutional,kondor2018covariant}.

Finally, of particular relevance to this paper is \cite{lahme1999karhunen} which explores similar themes (albeit in a somewhat different context) for the Karhunen-Lo\`{e}ve decomposition rather than multi-dimensional scaling. \cite{lahme1999karhunen} shows that by utilizing the intrinsic symmetries of a dataset, the computational burden of calculating the Karhunen-Lo\`{e}ve decomposition can be significantly reduced. Similar results will hold in many cases for MDS and we hope that in a future work we can explore the efficiency gains (as a function of the group and the metric) in greater depth.

The main contributions of this paper are the following
\begin{itemize}
    \item Explicit computations of the MDS eigendecomposition with Hamming distance on both the symmetric group and on the elementary abelian $2$-group (Section \ref{sect-hamming-distance}).
    \item Formula for the spectral decomposition of any bi-invariant metric on a group (Theorem \ref{theorem bi-invariant}).
\end{itemize}

This paper is organized as follows. In Section \ref{sect-multidimensional-scaling} we review the multidimensional scaling algorithm. Section \ref{sect-MDS-on-groups} contains the primary content of this paper. Here we consider what happens when MDS is applied to a dataset which is also a group. We begin this section with a quick overview of representation and character theory. In Section \ref{sect-hamming-distance} we explore the particular case where MDS is applied to a set of permutations with Hamming distance as the chosen metric. Two examples of using MDS on groups for data visualization are explored in \ref{sect-data-vis} followed by the conclusion in Section \ref{sect-conclusion}. 

\section{Multidimensional Scaling}
\label{sect-multidimensional-scaling}

Our main reference for the MDS algorithm is \cite{bibby1979multivariate}. Let $(X,d)$ be a finite metric space so that $X$ is a finite set of size $n$ and $d: X \times X \rightarrow \mathbb{R}_{\geq 0}$ is the metric on $X$ which encodes some notion of distance between points in $X$. Note that in this setting elements in $X$ need not come with intrinsic coordinates; only distances between elements. The input to MDS is the $n\times n$ pairwise distance matrix $\D$ defined by $(X,d)$ and the output is an embedding of $X$ into Euclidean space where the Euclidean distance (or more generally the pseudo-Euclidean distance) between points approximates $d$. 

For data visualization, one takes $k$ to be 2 or 3. Otherwise, the size of the embedding dimension $k$ is determined by the magnitudes of eigenvalues computed in the following way: define $\H:= \I_n - \frac{1}{n}\one \one^T$, where $\I_n$ is the $n \times n$ identity matrix and $\one$ is the $n\times 1$ vector of all ones. Define the double mean centered inner-product matrix 
\begin{equation*}
    \M:= -\frac{1}{2}\H (\D \circ \D) \H,
\end{equation*}
where $\circ$ denotes the Hadamard or element-wise product.

Compute the spectral decomposition: $\M = \F \Lam \F^H$.  A fundamental property of MDS is that $\M$ is positive semi-definite if and only if there exists a Euclidean configuration of points for which $D$ equals the Euclidean distances \cite[Theorem 14.2.1]{bibby1979multivariate}. In the case where $\D$ is not the matrix of a Euclidean configuration, there are two possibilities. 

In the classical algorithm, one discards any negative or zero eigenvalues of $\M$ and also the corresponding columns of $\F$. Define the $n\times k$ matrix $\X:=\widehat{\F} \widehat{\Lam}^{1/2}$, where the ``hat" indicates that the coordinates corresponding to eigenvalues less than or equal to $0$ have been removed. We follow the usual ordering convention ($\Lam$ is put in descending order). The rows of $\X$ give embedding coordinates to $X$ in Euclidean $k$-space. The Euclidean distance between rows $i$ and $j$ approximates the input distance $d_{ij}$ between points $x_i$ and $x_j$. MDS minimizes \emph{strain} on inner-products \cite[Section 14.4]{bibby1979multivariate}. If $k$ is the number of coordinate directions used, then the optimal MDS solution $\tilde{\M}$ has strain 
\begin{equation} \label{eqn-strain}
\text{tr}((\M-\tilde{\M})^2)=\sum_{i=k+1}^n\lambda_i^2,
\end{equation}
where $\{\lambda_i\}_{i = 1}^n$ are the eigenvalues from $\Lam$, i.e. the error is measured by the norm of the discarded eigenvalues. Perfect reconstruction on finite dimensional metric spaces is achieved by taking $k=n$. 

Our interest is in metric spaces which are decidedly non-Euclidean, so we anticipate information rich coordinate directions tagged by negative eigenvalues. One way \cite{pekalska2001generalized} to work with a non-Euclidean metric space is to keep both positive and negative eigenvalues and measure distances by a \emph{pseudo-Euclidean} computation. The process is simple and ought to be familiar to physicists working in Minkowski space. Suppose there are $p$ positive and $q$ negative eigenvalues with $p+q=k$ (as before, discard any zero eigenvalues). Arrange $\widehat{\Lam}$ into two corresponding blocks of size $p\times p$ and $q \times q$, and define the coordinates $\X:=\widehat{\F} \lvert \widehat{\Lam}\rvert ^{1/2} $. The pseudo-Euclidean distance computation breaks up into two parts - compute the Euclidean distance on the $p$ ``positive" coordinates and subtract off the Euclidean distance on the $q$ ``negative" coordinates. Just as in the Euclidean case, if we take $k=n$, we achieve an exact reconstruction of $\D$ by using the pseudo-distance computation on the rows of $\X$: $$d(x_i,x_j) = d_{\mathbb{C}^p}(\x_i,\x_j) - d_{\mathbb{C}^q}(\x_i,\x_j),$$ where $x_i$ denotes the coordinate free $i$th element of the set $X$, and $\x_i$ denotes its vector form from MDS. 

Finally, it's useful to understand MDS as a ``kernel method" \cite{scholkopf2001learning}, and think of the linear algebra outlined above as taking place in a Hilbert space of (square integrable) functions on $X$. In this language, the MDS kernel is the symmetric function $m: X \times X \rightarrow \mathbb{R}$ defined by $m(x_i,x_j)=\M_{ij}$.  Equivalent to the matrix $\M$ is the Hilbert-Schmidt operator $M:L^2(X) \rightarrow L^2(X)$, and the columns of $\F$ determine a collection $\{f_1, \ldots, f_k\}$ of eigen-functions for $M$. Then, $m(x_i,x_j)= \sum_{t=1}^n \lambda_t f_t(x_i)\overline{f_t(x_j)}$. 


\section{MDS on groups}
\label{sect-MDS-on-groups}

Suppose now that $G$ is a finite set which is also a group with group multiplication denoted by juxtiposition (i.e. if $g, h \in G$ then $gh$ is the product of $g$ and $h$).  If $d$ is a metric on $G$, we say that $d$ is \emph{left-invariant} to the action of $G$ if $d(g,h)=d(fg,fh)$ for all $f,g,h \in G$. We define a \emph{right-invariant metric} analogously. A metric which is both left and right invariant is called \emph{bi-invariant}.

For a given finite group $G$, consider the (non-centered) MDS kernel $m(g,h)=-\frac{1}{2}d^2(g,h)$ for a $G$-invariant metric $d$. Group invariance implies that $\M$ is a convolution matrix. Formally, if $\phi, \psi \in L^2(G)$ and $g,h \in G$, then convolution is defined by $$\phi \star \psi (h) :=\sum_{g \in G} \phi(hg^{-1})\psi(g).$$

Define $\mu(g) \in L^2(G)$ by $\mu(g):=m(g,e)$. The MDS operator $M:L^2(G) \rightarrow L^2(G)$ is defined for $g,h \in G$ by 
\begin{align*} 
M(\phi)(h)&= \sum_{g \in G} m(h,g)\phi(g)\\
&= \sum_{g \in G} \mu(hg^{-1})\phi(g)\\
&=\mu \star \phi (h).
\end{align*}
Every electrical engineer will recognize this as a generalization of ordinary convolution used to define linear time invariant operators. In our more general set-up, $\mu$ plays the role of the transfer function for band-pass filtering. Indeed, our goal in this paper is to compute the spectrum of frequencies amplified by the MDS operator for various groups. 

\subsection{Representations and Characters}

We briefly explain what is meant by ``frequency" for functions on an arbitrary finite group $G$. The reader is referred to \cite{Ser77,FH91} for rigorous accounts of the character and representation theory of groups. The short answer is that $L^2(G)$ decomposes into a set of mutually orthogonal subspaces, the so-called \emph{irreducible representations} specific to the group $G$. The presence of a frequency in a signal $\phi \in L^2(G)$ is determined by the amplitude of the projection coefficient onto the corresponding irreducible representation subspace. Schur's Lemma \cite[Proposition 4]{Ser77} guarantees that every linear $G$-equivariant operator (i.e. convolution operator) has a spectral decomposition whose eigenspaces are direct sums of irreducible representations.

In the classical case, when a signal is sampled at $n$ evenly spaced intervals, frequency information is determined by the discrete Fourier transform. In our language, the classical case corresponds to the cyclic group $G=C_n$, and the irreducible representations are tagged by integers $0, \ldots, n-1$ corresponding to the Fourier frequencies. Each irreducible representation determines a \emph{one dimensional subspace} in $L^2(C_n)$, the $k$th such subspace spanned by the function $f(m)=\exp{\frac{2\pi i mk}{n}} \in L^2(C_n)$. The main difference for an arbitrary group $G$ is that a single frequency may account for a subspace in $L^2(G)$ of dimension greater than $1$. In fact, such a frequency always exists for any group which is non-abelian. 

Informally, a \emph{representation} of $G$ assigns to each element of $G$ an invertible matrix so that the group multiplication law of $G$ is realized by matrix multiplication. Formally, an $n$-dimensional representation of $G$ is a pair $(V, \rho)$ where $V$ is a (for our purposes, complex) $n$-dimensional vector space and $\rho$ is a group homomorphism from $G$ to the general linear group on $V$, which is the group of all invertible linear transformations on $V$ with group law given by composition of transformations. When the representation $(\rho, V)$ is understood from context we may, for convenience, omit the $\rho$ from our notation e.g. if $g \in G, v \in V$, we write $g \cdot v$ to indicate the application of transformation $g$ to vector $v$, instead of $\rho(g)v$. 

A representation $V$ is \emph{reducible} if there exists a non-trivial, proper subspace $W \subset V$ such that $W$ is preserved by all transformations of $G$ i.e. for all $w \in W$ and all $g \in G$, $\rho(g) w \in W$. If $V$ is not reducible, it is called irreducible. Maschke's theorem \cite[Theorem 2]{Ser77} guarantees that every complex representation $V$ of a finite group $G$ decomposes into a unique set of irreducible representations, which comprise a decomposition of $V$ into orthogonal subspaces.  

Note that $L^2(G)$ is a representation of $G$: it is a vector space of dimension $n=|G|$ and each element of $G$ acts as a linear transformation on $L^2(G)$ by the rule $g \cdot \phi(h):= \phi(g^{-1}h)$.  

Next, we investigate how MDS filters frequencies in $L^2(G)$ for the symmetric group. To do so will be an exercise in the linear algebra of \emph{characters}: Associated to any representation $(V,\rho)$ is the character $\chi_\rho$, which is an element of $L^2(G)$ defined by $\chi_\rho(g)=Tr(g)$, the trace of the linear transformation. It turns out that characters uniquely determine irreducible representations. 

If $(V,\rho)$ and $(V',\rho')$ are distinct irreducible representations, then the characters are orthogonal under the $L^2$ inner product:
\begin{align*}
\langle \chi_\rho, \chi_\rho' \rangle &:= \frac{1}{|G|} \sum_{g \in G} \chi_\rho(g)\overline{\chi_{\rho'}}(g)\\
&=\left\lbrace \begin{array}{cc}0 & (V,\rho) \not\cong (V',\rho')\\ 1 & (V,\rho) \cong (V',\rho') \end{array} \right..
\end{align*} 

We now state the fundamental relationship between characters, bi-invariant metrics, and multi-dimensional scaling. 

\begin{theorem} \label{theorem bi-invariant}
Let $m$ be the MDS kernel and $\mu(g):=m(g,e)$. If $d$ is a bi-invariant metric on $G$, then 
\begin{itemize}
\item[i.] $$\mu(g)=\sum_{i=1}^k \sigma_i \chi_{\rho_i}(g),$$
where $\sigma_i \in \mathbb{\mathbb{R}}$ and each $\chi_{\rho_i}$ is the character of an irreducible representation $\rho_i$ of $G$. 
\item[ii.] Each irreducible $(V_i, \rho_i)$ which appears in the sum determines an eigenspace of the MDS operator $M$. If $d_i:=\dim(V_i)$, then the eigenvalue associated to $V_i \subset L^2(G)$ is given by $\lambda_i = \frac{|G|}{d_i}\sigma_i$. 
\end{itemize}
Equivalently, the spectral decomposition of the MDS matrix is $\M=\sum_{i=1}^k \lambda_i \V_i \V_i^H$.
\end{theorem}

The Theorem gives us a concrete way of computing which frequencies are filtered by the MDS operator for a given bi-invariant metric. 

Also note that the trivial representation, whose character is $\chi(g)=1$ for all $g \in G$, necessarily appears as an eigenfunction of the MDS operator $M$. The double mean centering step in the MDS algorithm will project the trivial character to $0$ and leave the other eigenfunctions fixed (by the orthogonality of characters). Then, for simplicity, we use the non-centered MDS kernel $m(g,h):=-\frac{1}{2} d^2(g,h)$ for our computations and simply remember to project away from the trivial representation. 

The theorem follows from two facts. (1) If $d$ is a bi-invariant function in $L^2(G)$, then for all $g,h \in G$, $\mu(hgh^{-1})= \mu(g)$. It is said that $\mu$ is a \emph{class function} on $G$. (2) The characters of the irreducible representations form an orthonormal basis for all class functions on $G$ \cite[Proposition 2.30]{FH91}. We leave it to the reader to verify the formula for the eigenvalues, which follow from these two facts and that $M$ is convlution with $\mu$. 

\section{MDS with Hamming distance}
\label{sect-hamming-distance}
In this section we derive the eigendecomposition of MDS under the Hamming distance on two frequently used groups.

\subsection{Binary data and the Hamming metric} 
\label{ex-binarydataham}
Let $C_2$ be the cyclic group of order $2$ and let $G=(C_2)^k$ be the product of $k$ copies of $C_2$. The order of $G$ is $n=2^k$ and elements of $G$ can be represented by length $k$ strings of $0$'s and $1$'s. The \emph{Hamming distance} on $G$ counts the number of positions at which two binary strings differ. For example, $d_H(00,11)=2$. $d_H$ is a $G$ bi-invariant metric, and by Theorem \ref{theorem bi-invariant} we are guaranteed that the MDS kernel can be written as a sum of irreducible characters of $G$. Moreover, since $G$ is abelian, its irreducible representations are all $1$ dimensional. 

Represent an element in $G$ by $g=(n_1, \ldots, n_k),\; n_i \in \{0,1\}$. It's well-known that the irreducible characters are indexed by elements $\mathcal{S}$ of the power set on $\{1, \ldots, k\}$. For a non-empty $\mathcal{S}$, define $$\chi_\mathcal{S}(n_1, \ldots, n_k) := (-1)^{\sum_{s \in \mathcal{S}} n_s},$$ and define the character corresponding to the empty set to be the one's function on $G$.  In discrete signal processing, these are called the Walsh functions.  What group theorists call the character table is exactly the same as the Walsh-Hadamard matrix. If $k=2$, the character table is:

\vspace{2mm}
\begin{equation*}
\begin{array}{c|c|c|c|c}

  & \mathbf{00} & \mathbf{10} & \mathbf{01} & \mathbf{11} \\ 
 \hline 
 \chi_{t^2} & 1 & 1 & 1 & 1 \\ 
 \hline 
 \chi_{ta} & 1 & 1 & -1 & -1 \\ 
 \hline 
 \chi_{at} & 1 & -1 & 1 & -1 \\ 
 \hline 
 \chi_{a^2} & 1 & -1 & -1 & 1 \\ 

 \end{array} \;\;,
\end{equation*}
where $t$ and $a$ refer to \emph{trivial} and \emph{alternating} which is the language used by group theorists. 

Now, Theorem \ref{theorem bi-invariant} suggests that we should decompose $\mu(g):=m(g,e)$ into irreducible characters. To begin, note that the distance between any string $g=(n_1, \ldots, n_k)$ and the identity string $e=(0, \ldots, 0)$ counts the number of ones in string $g$. We may decompose this function as a sum of irreducible characters: $$d_H(g,e) = \frac{1}{2}(k- \sum_{i=1}^k (-1)^{n_i}).$$

Then, the MDS kernel is given by:
\begin{align*}
\mu(g)&:=-\frac{1}{2}\left( \frac{1}{2}(k- \sum_{i=1}^k (-1)^{n_i}\right)^2\\
&=-\frac{1}{8}\left[ k^2 -2k \sum_1^k (-1)^{n_i} +\left(\sum_1^k(-1)^{n_i} \right)^2 \right] \\
&=-\frac{1}{8}\left( k^2 +k \right) +\frac{k}{4} \sum_{1}^k(-1)^{n_i} -\frac{1}{4} \sum_{1 \leq i < j \leq k} (-1)^{n_i+n_j}.
\end{align*}

Now we simply read off the eigendecomposition of the MDS operator. In particular, the appearance of a character in the sum gives an eigenfunction, and the coefficients give eigenvalues (after multiplication by $|G|=2^k$). Remember also that the MDS algorithm calls for projection away from the trivial representation, and so we discard the translation term out front. We summarize the eigendecomposition in Figure \ref{fig-hamming-dist-table-z2}.

\begin{figure}
\begin{center}
$\begin{array}{|c|c|c|}
\hline
\text{character} & \text{eigenvalue} & \text{num principal directions} \\ 
\hline
\chi_{t^{k-1}a} & \lambda_1=2^{k-2}\cdot k & k\\ 
\chi_{t^{k-2}a^2} & \lambda_2=-2^{k-2} & \binom{k}{2} \\
\text{else} &0 & 2^k-(k+\binom{k}{2})\\
\hline
\end{array} $
\end{center}
\caption{\label{fig-hamming-dist-table-z2} The MDS eigenvalues and number of principal directions associated with Hamming distance on $G=C_2^k$.}
\end{figure}

A few observations are in order. First, the computation reveals low dimensional structure, as the distance matrix itself is $2^k \times 2^k$, yet the rank is only $k+\binom{k}{2}$. 

Next, using strain \eqref{eqn-strain} as our measurement of projection error, principal directions corresponding to $\lambda_1$ contribute more than those corresponing to $\lambda_2$, and any directions with the same eigenvalue are equally strong.

Finally, note that the first $k$ coordinates are tagged with a positive eigenvalue and the last $\binom{k}{2}$ are tagged with a negative eigenvalue. This gives a measure of the extent to which the metric space $(C_2^k, d_h)$ is Euclidean. Formally, this means that we use a pseudo-Euclidean inner-product to make geometric measurements. 

\subsection{Hamming metric on the symmetric group}
In this section we explore another type of Hamming metric, only this time on the symmetric group $G:=S_n$.  As a prerequisite for this section, the reader should understand what is meant by the ``standard representation" of $S_n$. We refer the reader to \cite{Sag01,FH91} for more details. 

Of fundamental importance to the representation theory of $S_n$ is that there is one irreducible representation for each \emph{integer partition} of $n$. Compare this to the discrete Fourier case where frequencies are tagged by integers $0, \ldots ,n-1$, or, as in the last section, to the case $C_2^k$, where frequencies are tagged by elements of the power set on $\{1,\ldots, k\}$. 

We use square brackets to denote partitions e.g. If $n=3$ then $2+1=3$ is denoted $[2,1]$, and the trivial partition $3=3$ is denoted $[3]$, etc. We denote the character associated to the irreducible $[2,1]$ by $\chi_{2,1}$.  

The Hamming distance $d_H$ between two permutations counts the number of places where the two permutations differ e.g. Using the cycle notation for permutations we have $d_H( (12), (123))= 2$, since the two permutations agree only on $1 \mapsto 2$. It's straight-forward to check that this metric is bi-invariant on $S_n$. 

As in the last section, our goal is to produce the MDS eigendecomposition by decomposing $\mu(g)$ into characters. The Hamming distance between permutation $g$ and the identity permutation $e$ is given by the difference between $n$ and the number of fixed points of the permutation $g$.  i.e. 
\begin{equation*}
d(g,e) = n- \#f.p.(g).
\end{equation*}

We can write this in terms of the characters of irreducible representations: $$d(g,e)=(n-1)\chi_{n}(g)-\chi_{n-1,1}(g),$$ where $\chi_{n}$ is the character of the trivial representation (which equals $1$ on all elements of $S_n$) and $\chi_{n-1,1}$ is the character of the standard representation. Squaring this expression gives the MDS kernel $m$:
\begin{align*}
-2\mu(g)&=\left( (n-1)\chi_{n}(g)-\chi_{n-1,1}(g)\right)^2\\
&=(n-1)^2\chi_{n}(g)-2(n-1)\chi_{n-1,1}(g)+\chi_{n-1,1}^2(g)
\end{align*}

All that remains is to decompose the squared character $\chi_{n-1,1}^2$ into a sum of irreducible characters, which we can do since irreducible characters form an orthogonal basis for class functions. This is accomplished using the fact that the square of a character corresponds to the tensor product of the underlying representation space, and then extracting the Kronecker coefficients. In our case, the formula is simply \cite{FH91}:
\begin{equation*}
    \chi_{n-1,1}^2=\chi_{n}+\chi_{n-1,1}+\chi_{n-2,1,1}+\chi_{n-2,2}.
\end{equation*}
Table \ref{fig-hamming-dist-table} then summarizes the MDS embedding of Hamming distance in terms of its decomposition into irreducibles. Note that the formula for each eigenvalue relies on the dimension of the corresponding representation, which may be computed using the ``hook-length" formula \cite{FH91}. 

Here, the energy is highly concentrated in only three subspaces. Moreover, while the group has order $n!$ the rank of the MDS matrix is on the order of $n^4$. The Euclidean coordinates of the metric are picked up by the standard representation, which is also the dominant representation, whereas the pseudo-Euclidean coordinates are given by the subspaces of $\chi_{n-2,1,1}$ and $\chi_{n-2,2}$.

\begin{figure}
\begin{center}
$\begin{array}{|c|c|c|}
\hline
\text{character} & \text{eigenvalue} & \text{num principal directions} \\ 
\hline
\chi_{n-1,1} & \frac{(2n-3)n!}{2n-2} & (n-1)^2 \\ \hline
\chi_{n-2,1,1} & \frac{-n!}{(n-1)(n-2)} & \left(\frac{1}{2}(n-1)(n-2)\right)^2\\ \hline 
\chi_{n-2,2} & \frac{-n!}{n(n-3)} & \left(\frac{1}{2}n(n-3)\right)^2 \\
\hline
\end{array} $
\end{center}
\caption{\label{fig-hamming-dist-table} The MDS eigenvalues and number of principal directions associated with Hamming distance decomposed in terms of irreducible representations of the symmetric group $S_n$. Note that only a minority of the irreducible representations of $S_n$ (indexed by all partitions of $n$) have support here.}
\end{figure}

\section{Data Visualization Examples}
\label{sect-data-vis}

In this section we apply MDS to two datasets that take values in a group and to which we apply a bi-invariant metric.

\begin{example}
 The first dataset is a set of rankings from the American Psychological Association (APA) presidential election in 1980. This dataset can be found in \cite[Chapter 5B]{Dia88}. It consists of 5,738 full rankings of $5$ candidates. The original dataset included partial rankings, but we have omitted these. Of course we can interpret these full rankings as permutations by choosing an initial order of the candidates. A given ranking corresponds to the permutation that takes the original order to the order given by the ranking.

 
 In Figure \ref{fig-APA} we show the MDS approximation of the permutations in this dataset in $\R^{3}$ with respect to Hamming distance (without scaling). We use the size of points in the scatterplot to indicate the frequency of a particular permutation and use color to indicate a fourth coordinate (also taken from the block of standard representations). 
 
The second dataset is the SUSHI preference dataset \cite{kamishima2005supervised} which contains $5000$ full rankings of $10$ types of sushi. Note that whereas in the APA election dataset there are more data points than there are permutations ($5,738$ vs $5! = 120$), in this dataset there are far more possible permutations ($5,000$ vs $10! = 3,628,800$) than there are data points. In Figure \ref{fig-sushi} we show a visualization of all these rankings in $\mathbb{R}^3$ using MDS. As seen above, despite the fact that distances on permutations have the capacity to be quite high dimensional, by understanding what symmetric representations actually contribute information to the Euclidean embedding, we can directly project into these representations because applying the MDS dimensionality reduction algorithm. 
 
 \begin{figure}
\begin{center}
\includegraphics[width=8cm]{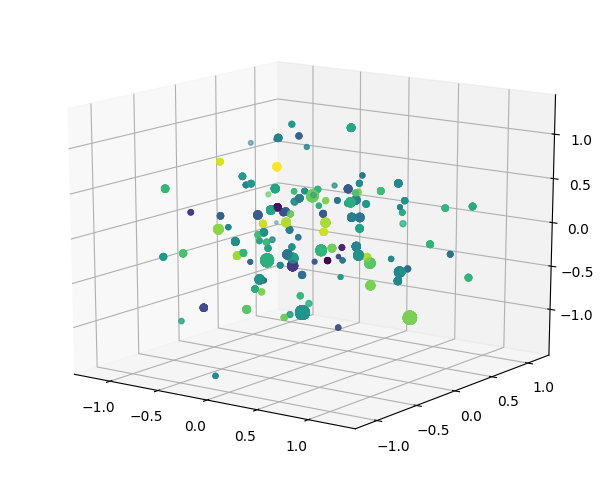}
    \caption{A visualization of rankings of the American Psychological Association presidential election from 1980 \cite[Chapter 5B]{Dia88} using MDS.}
    \label{fig-APA}
\end{center}
\end{figure}

\begin{figure}
\begin{center}
\includegraphics[width=8cm,height=6cm]{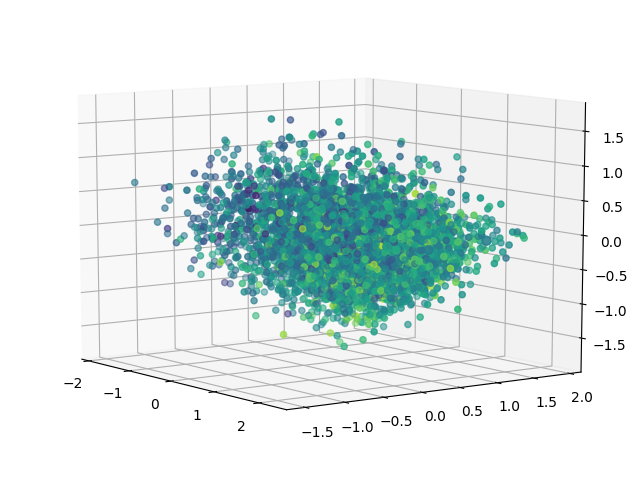}
   \caption{A visualization of the SUSHI dataset \cite{kamishima2005supervised} using MDS.}
    \label{fig-sushi}
\end{center}
\end{figure}
 

\end{example}

\section{Conclusion}
\label{sect-conclusion}
 In this paper we have shown how unstructured data can be analyzed and synthesized using the general notion of frequency on a group and the MDS algorithm. We have seen how the principal directions extracted from MDS are given geometric meaning as irreducible representations, and how each representation contributes to the pseudo-Euclidean structure of the group metric space.  
 
In practical terms, the theory and examples presented here may be used for dimensionality reduction. In a future work we plan to more closely investigate the efficiency gains brought by group theory considerations as well as analysis of other commonly encountered groups and metrics.

Since this work lies in the intersection of metric geometry, group theory, and data analysis, we hope this paper is useful for a wide range of audiences.

\bibliographystyle{amsplain}


%
%
%
%
\bibliography{MDSRefs}

\end{document}